\documentclass{article}

\usepackage{xcolor}

\newcommand{\etal}{\textit{et al.}}

\usepackage{tikz,subcaption}
\usetikzlibrary{spy,backgrounds,positioning,shapes,arrows}
\usepackage{xcolor}
\usepackage{amsmath}
\usepackage[ruled]{algorithm2e}
\usepackage{nicefrac}
\usepackage{indentfirst}
\usepackage{tabularx}
\usepackage{color}
\usepackage{colortbl}
\usepackage{graphicx}
\usepackage{paralist}
\usepackage{lipsum}
\usepackage{wrapfig}
\usepackage{multirow}
\usepackage{textcomp}
\usepackage{gensymb} 
\usepackage{bm} 
\usepackage{lmodern}

\definecolor{TransportFlow}{RGB}{255, 0, 0}
\definecolor{DPPO}{RGB}{220, 80, 160}
\definecolor{ReinFlow}{RGB}{0, 100, 200}
\definecolor{FPO}{RGB}{0, 100, 200}

\usepackage{natbib}
\usepackage[preprint]{neurips_2025}

\usepackage[utf8]{inputenc} 
\usepackage[T1]{fontenc}    
\usepackage{hyperref}       
\usepackage{url}            
\usepackage{booktabs}       
\usepackage{amsfonts}       
\usepackage{nicefrac}       
\usepackage{microtype}      
\usepackage{xcolor}         

\title{Reinforcement Learning for Flow-Matching Policies with Density Transport}

\author{%
  Boshu Lei \\
  University of Pennsylvania \\
  Philadelphia, PA 19103 \\
  \texttt{leiboshu@seas.upenn.edu} \\
  \And
  Kostas Daniilidis \\
  University of Pennsylvania \\
  Philadelphia, PA 19103 \\
  \texttt{kostas@cis.upenn.edu} \\
  \And
  Antonio Loquercio \\
  University of Pennsylvania \\
  Philadelphia, PA 19103 \\
  \texttt{aloque@seas.upenn.edu} \\
}

\begin{document}

\maketitle

\begin{abstract}
We present an online reinforcement learning (RL) algorithm for fine-tuning flow-matching policies in continuous-control problems.
Our key insight is to view RL-based policy improvement as a transport of action densities towards regions of high reward, which naturally aligns with the transport formulation of flow matching models.
Prior methods either approximate the current or optimal policy distribution or resort to distillation, which introduces biased gradients or sacrifices multimodal modeling capacity.
In contrast, our approach for RL with Density Transport, which we name \emph{RLDT}, constructs a transport field from a maximum-entropy RL objective using Stein Variational Gradient Descent (SVGD). Then, it finetunes a pretrained flow matching policy to align with this field.
Training with this alignment objective is nontrivial because flow-matching policies generate actions via a multi-step process, making direct gradient-based optimization challenging. 
To overcome this challenge and stabilize training, we approximate policy actions from intermediate denoising steps via expected-target estimation. This allows the transport-field update to propagate into the network parameters without unstable backpropagation through time.
Experimental results demonstrate that RLDT outperforms competitive baselines in reward quality and convergence speed. This performance holds across diverse continuous-control tasks, encompassing both dense and sparse rewards, as well as state- and vision-based long-horizon robot manipulation. The project webpage is \href{https://rpfey.github.io/rldt/}{https://rpfey.github.io/rldt/}.
\end{abstract}

\section{Introduction} 
\label{sec:intro}

Vision-Language-Action (VLA) models~\cite{kevin2024pi0,wu2026pragmatic,shukor2025smolvla,moo2024openvla} have demonstrated strong capacity to encode rich priors over state-action distributions, making them well-suited as foundations for downstream task adaptation. 
However, the learned prior, generally referred to as \emph{base policy}, may not be optimal for a specific use case. For example, the trajectories in the training dataset could be suboptimal, or the policy may fail on scenarios outside its training distribution.
When a tractable approximation of the task-specific optimal policy is available, it can be used to steer action generation without modifying the model parameters~\cite{ye2025steeringdiffusionpolicy,du2025dynaguidesteeringdiffusionpolices,fang2024diffusion,psenka2024learning}.
When no such approximation is accessible, however, reinforcement learning offers a principled way to fine-tune the base policy on the target task.

The base policy admits a multitude of instantiations, e.g., diffusion~\cite{barreiros2026careful}, flow-matching~\cite{yaron2024flowmatchingtutorial}, or advanced tokenization~\cite{pertsch2025fast}. In this work, we focus on flow-matching policies, following their successful adoption in state-of-the-art models for robotics~\cite{wu2026pragmatic,pi05,shukor2025smolvla}. Training a flow-matching policy with reinforcement learning raises two main challenges: (1) we don't have samples from the task-specific optimal policy; (2) estimating the policy log-likelihood is very expensive due to the multi-step denoising process required to generate actions.


Prior work has addressed these issues in a variety of ways. One popular approach is optimizing lower bounds on the policy log-likelihood~\cite{yi2026fpo++,mcallister2026fpo,black_dppo}, which, however, leads to biased policy gradients. Other methods ignore the gradients at intermediate denoising steps~\cite{rendiffusion}, or solve the reverse ODE to obtain a per-step target~\cite{qiyang2026qam}. Other works introduce changes to the policy structure by distilling a multi-step policy into an actor requiring only a few steps~\cite{park2025flowqlearning, zhang2025reinflowfinetuningflowmatching}, which simplifies the RL problem and enables backpropagation through the denoising process. Similarly, other works model flow-matching as a recurrent neural network and gate the velocity to control the gradient magnitude~\cite{zhang2026sacflow}. While effective in mitigating the problem of vanishing or exploding gradients, these methods constrain the expressivity of the trained policy, potentially leading to suboptimal performance.

\begin{figure}[t]
    \centering
    \includegraphics[width=0.95\linewidth]{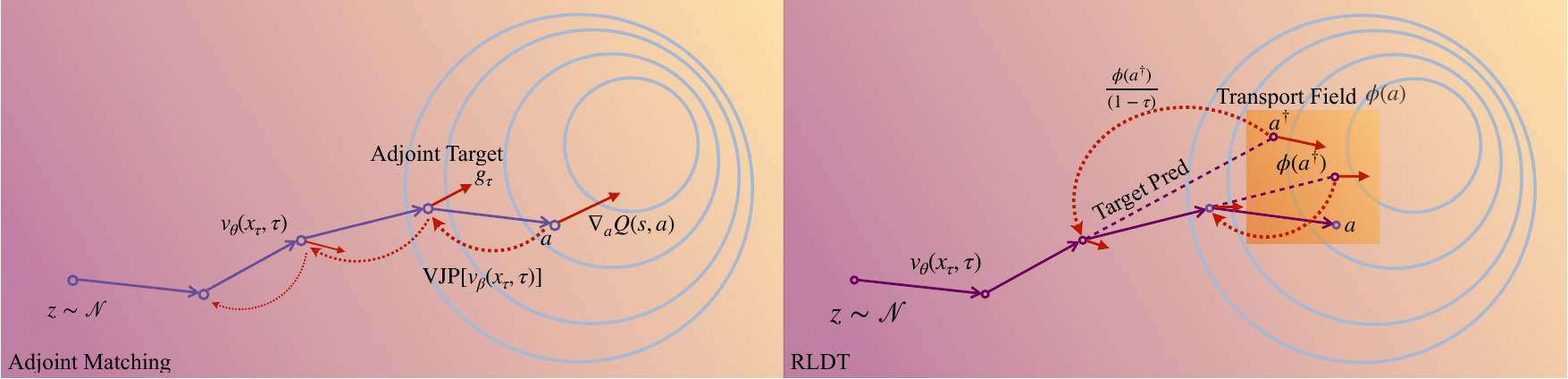}
    \caption{\textbf{RLDT Pipeline} Left: QAM~\cite{qiyang2026qam} uses the Q gradient at the final sample and the Vector Jacobian Product (VJP) of a reference policy $\pi_\beta$ to compute intermediate adjoint targets $g_\tau$ to update $\pi_\theta$. Right: RLDT constructs a transport field $\phi(a)$ on the action manifold and uses the expected posterior $a^\dagger$ to compute targets for intermediate denoising steps of $\pi_\theta$.}
    \label{fig:pipeline}
    \vspace{-2ex}
\end{figure}

In this paper, we propose Reinforcement Learning with Density Transport (\textbf{RLDT}), which recasts the RL policy improvement step as a probability transport problem \emph{on the action manifold}. As illustrated in Fig.~\ref{fig:pipeline}, RLDT aligns a flow-matching field to a transport field $\phi(a): \mathbb{R}^d \rightarrow \mathbb{R}^d$ that drives action probabilities toward regions of higher expected reward. This formulation offers two key advantages over prior work: (i) it circumvents the exact computation of action log-likelihoods, and (ii) it enables direct gradient propagation through intermediate denoising steps without requiring structural policy changes, solving a reverse ODE, or proxy objectives.

We employ Stein Variational Gradient Descent (SVGD)~\cite{liu2016steinvaritaionalgradientdescent} to compute the transport vector field $\phi(a)$. This field is designed to move samples across the action manifold toward the density of the optimal policy, as defined by a maximum-entropy RL objective. To mitigate training instabilities inherent in multi-step generation, we estimate the policy's actions from intermediate steps of the ODE using expected target prediction~\cite{yaron2024flowmatchingtutorial}. Because these predicted targets reside on the action manifold, we can compute the SVGD update directly on them and back-propagate the resulting gradients into the flow-matching backbone.

We evaluate RLDT across three benchmark settings of increasing complexity: 
continuous control with dense rewards (OpenAI Gym), long-horizon multi-stage robot manipulation with sparse task rewards and state-based information (Furniture-Bench~\cite{heo2023furniturebench}), and vision-based robot manipulation with sparse rewards (Robomimic~\cite{robomimic2021}). While most prior research on RL for diffusion or flow models focuses on state-based, short-horizon tasks, to the best of our knowledge, only one concurrent study~\cite{rendiffusion} has conducted a full evaluation on similarly diverse and high-dimensional robot control tasks. Our empirical findings demonstrate that RLDT consistently outperforms a variety of baselines across all settings. 
We further show that, in contrast to prior methods that generate targets for intermediate denoising steps, our method produces well-conditioned gradient magnitudes.

Overall, our contributions can be summarized as follows:
\begin{compactitem}
    \item We formulate a novel RL algorithm for finetuning flow-matching policies by modeling policy improvement as a transport of action probabilities towards regions of high reward. Our formulation bypasses density estimation and backpropagation through time, leading to better performance than baselines and stable training.
    \item We tackle the problem of backpropagation through the denoising process of flow-matching policies by estimating the expected policy action from intermediate denoising steps. Since the transport field is defined on the action manifold, this prediction enables gradient back-propagation through the flow-matching backbone over the entire denoising process.
\end{compactitem}


\section{Related Work}
\label{sec:related-works}

\textbf{On-Policy RL for Diffusion/Flow Policies.} Methods in this category generally consist of on-policy algorithms inspired by PPO \cite{schulman2017ppo}. The primary challenge in adapting these for flow policies lies in estimating the policy log-likelihood to derive gradient updates. Because exact computation is computationally expensive for flow models, existing works typically utilize lower-bound approximations. For instance, DPPO \cite{rendiffusion} computes the log-likelihood of all denoising latent variables as a lower bound of the marginal probability. To add stochasticity to the sampling process for exploration and gradient estimation, prior work transforms the flow ODE to an equivalent SDE~\cite{liu2025flow} or inject learnable noise~\cite{sun2025f5r}. This strategy has been applied to continuous control problems, which is the focus of this work, in ReinFlow \cite{zhang2025reinflowfinetuningflowmatching} and SACFlow \cite{zhang2026sacflow}. These methods, however, remain dependent on density-based objectives. FPO \cite{mcallister2026fpo} utilizes the Evidence Lower Bound (ELBO) \cite{kingma2023understandingdiffusionobjectiveastheelbo} to optimize the policy; however, when the advantage is negative, minimizing a lower bound does not strictly guarantee a reduction in the true policy log-likelihood. To mitigate this, FPO++ \cite{yi2026fpo++} employs a particle ensemble and an asymmetric trust region to dampen gradients from negative advantages. In contrast to these methods, our work uses a value-based framework that bypasses the need to approximate the policy's log-likelihood entirely.

\textbf{Off-Policy RL for Diffusion/Flow Policies.} Methods in this category are primarily distinguished by how they leverage the critic to guide policy optimization. One approach employs the critic for rejection sampling, refining the policy by filtering for high-value actions~\cite{hansen2023idql,dong2025expo,pfrommer2025reinforcement, OTPR}, potentially learning a reward surrogate to improve sample efficiency~\cite{pfrommer2025reinforcement}. 
Notably, Sun~\etal~\cite{OTPR} views the Q value as a transport cost to build the optimal transport map from states to actions by selecting higher Q value actions from sampled actions.
However, this is generally restricted to minor deviations from the base policy, especially given a high-dimensional action space.
Alternatively, some methods backpropagate the critic's gradient through the entire multi-step denoising chain~\cite{wang2016learning,ding2023consistency,zhang2024entropy}. This approach often suffers from instability, particularly as the number of denoising steps increases~\cite{park2025flowqlearning}. 
To mitigate this, distillation techniques have been proposed to compress flow-matching policies into few-step models~\cite{park2025flowqlearning,espinosa2025scaling,chen2025one}, though this frequently comes at the cost of the model's expressive power. Most relevant to our work are methods that bypass backpropagation through the denoising process by constructing step-wise supervision targets. While this introduces potential bias, the central challenge lies in designing targets that maintain stability without sacrificing optimality. Existing strategies address this by approximating the optimal policy~\cite{lu2023contrastive,fang2024diffusion,khan2026flow,alles2025flowq} or substituting the critic’s gradient with its scalar value~\cite{ma2025efficient,koirala2025flow,zhang2024entropy,haitong2025efficientonlinereinforcementlearningfordiffusionpolicy,psenka2024learning}, potentially with additional losses to stabilize learning, e.g., entropy~\cite{gao2026flow}. 
The method most closely related to ours utilizes adjoint matching to create step-wise objectives~\cite{qiyang2026qam}. Similarly to ours, this provides theoretical convergence guarantees. However, we observe that its gradients are less well-balanced over intermediate steps than ours. 

\textbf{Optimal Transport for Flow Models} Prior work has shown that optimal transport can be used to match noise and data pairs to improve training efficiency~\cite{mousavi2025flow}. More related to our work, Kong~\etal~\cite{kongcomposite25} use optimal transport for finetuning a flow policy with RL when the dynamics of the target task is different to the one used to train the base policy. Specifically, they use OT for filtering rollouts and biasing exploration towards high-uncertainty states. Our transport perspective is used to build the RL objective and not only for filtering.


\textbf{RL via Residuals or Guidance}. These methods preserve the pretrained base policy while leveraging a few learnable parameters to modulate the action distribution on the target task. One approach involves training a lightweight residual actor, trained with RL, operating on the base policy's output~\cite{pi07,ankile2024fromimitationtorefinement,wang2026omnixtreme}. Alternatively, some methods optimize the generation process itself by using a Q-network to learn a steering noise or latent shift, which biases the denoising trajectory toward high-reward regions of the action space~\cite{ye2025steeringdiffusionpolicy,su2026rfs}. Guidance can also be achieved by using sampling-based optimizers like MPPI~\cite{wang2024residual}. While these strategies maintain the robustness of the foundation model, their reliance on a frozen base can be a bottleneck when the base policy's initial coverage of the target task is poor.

\section{Method}
\label{sec:method}
Our core insight is to cast RL as a transport map $T = \mathrm{id} + \epsilon\phi$, where $\mathrm{id}$ is the identity mapping and  $\phi$ is the transport velocity field. This framework naturally fits flow-matching policies, which learn a time-dependent transport field from a simple prior distribution, e.g., white noise, to the target action distribution.

Following standard convention, we model reinforcement learning as a Markov Decision Process (MDP) defined by the tuple $(\mathcal{S}, \mathcal{A}, p_s, r)$, where the state space $\mathcal{S}$ and action space $\mathcal{A}$ are continuous. 
The environment's (unknown) dynamics are characterized by the state transition probability density $p_s(s_{t+1} | s_{t}, a_{t})$, where $p_s: \mathcal{S} \times \mathcal{S} \times \mathcal{A} \rightarrow [0, \infty)$. This function represents the probability density of the next state $s_{t+1} \in \mathcal{S}$ given the current state $s_{t} \in \mathcal{S}$ and action $a_{t} \in \mathcal{A}$. At each time step, the environment yields a bounded scalar reward $r(s_t,a_t) \in [r_{\min}, r_{\max}]$.



\subsection{Flow Matching Policies}
\label{sec:flow_match}
Flow Matching~\cite{Lipman2022FlowMatching} learns a velocity field $v_\theta(x_{\tau}, \tau)$ that generates a probability path  $p_\tau(x)$, $\tau\in[0,1]$, between an initial distribution $q(x)$ and a target distribution $p(x)$, such that $p_0(x) = q(x)$ and $p_1(x) = p(x)$. 
The velocity field $v_\theta(x_\tau, \tau)$ is used to construct a time-dependent diffeomorphic map $\psi_\tau(x): [0,1] \times \mathbb{R}^n \rightarrow \mathbb{R}^n$ defined via the ordinary differential equation:
\begin{equation}
    \begin{aligned}
        \frac{d \psi_\tau(x)}{d\tau} &= v_\theta(\psi_\tau(x), \tau) \\
        \psi_0(x) &= x
    \end{aligned}
    \label{eq:flow-ode}
\end{equation}
The solution of this ODE maps a sample from the starting distribution $q(x)$ to the target distribution $p(x)$. Following~\cite{yaron2024flowmatchingtutorial}, we train the velocity field $v_\theta(\psi_\tau(x), \tau)$ with the following objective:
\begin{equation}
    \min \limits_{\theta} \mathbb{E}_{\tau \sim \mathcal{U}[0, 1], x_1 \sim p(x), x_0 \sim q(x)} \Vert v_\theta(\psi_\tau(x), \tau) - (x_1 - x_0) \Vert^2.
    \label{eq:flow_train}
\end{equation}
In practice, at inference time we sample from $x_0\sim q(x)$ and iteratively apply the velocity field for a fixed number of steps to obtain a sample $x_1\sim p(x)$.

For a flow-matching policy, the random variables $x$ are actions $a_t$, and the velocity field is conditioned on the state $s_t$. The initial distribution $q(a_t) = \mathcal{N}(0, I)$ and the target distribution $p(a_t)=\pi^*(a_t|s_t)$ is the optimal policy. To distinguish between the time $t$ in the MDP and $\tau$ in the ODE, we refer to an intermediate step at $\tau$ of a flow matching policy as $a_{t,\tau}$.
An important advantage of the optimization in Eq.~\eqref{eq:flow_train} is that it enables training a policy only from samples of $\pi^*(a_t|s_t)$ (generally referred to as expert demonstrations).

From Eq. \eqref{eq:flow-ode}, we can see that a flow-matching policy effectively learns a vector field over the action manifold. In the following section, we demonstrate how to align this field to an optimal policy field for a maximum-entropy RL objective. This alignment yields a tractable, off-policy reinforcement learning algorithm.



\subsection{Maximum Entropy Reinforcement Learning for Flow Matching Policies}
\label{sec:transportation}


Following the maximum-entropy formulation~\cite{ziebart2010maximumcausalentropy}, we define the RL optimization objective as:
\begin{equation}
    \pi^* = \arg\max_{\pi} \mathbb{E}_{(s_0,a_0, \dots, s_{T-1}, a_{T-1}) \sim \rho_\pi} \left[ \sum_{t=0}^{T-1} \gamma^t \left( r(s_t, a_t) + \kappa \mathcal{H}(\pi(\cdot \mid s_t)) \right) \right]
    \label{eq:max_ent}
\end{equation}
where $\rho_\pi$ is trajectory distribution induced by a policy $\pi$ and $\kappa$ is a constant to control entropy regularization. The solution to Eq.~\eqref{eq:max_ent} can be achieved by iterative soft Q iteration~\cite{Haarnoja2017reinforcementlearningwithdeepenergybasedpolicy}:
\begin{equation}
    \begin{aligned}
        Q(s_t, a_t) &\leftarrow r(s_t,a_t) + \gamma \mathbb{E}_{s_{t + 1}} [V(s_{t + 1})] \\
        V(s_t) &\leftarrow \kappa \log \int_{\mathcal{A}} \exp \left (  \frac{1}{\kappa} Q (s_t, a')  \right ) da' \\
        \pi^* (a_t|s_t)  &= \frac{\exp(Q(s_t, a_t) / \kappa)}{Z},
    \end{aligned}
\end{equation}
where $Z$ is the normalization term. Similarly to~\cite{Haarnoja2017reinforcementlearningwithdeepenergybasedpolicy}, we train a stochastic policy $\pi_\theta = \pi(a_t|s_t;\theta)$ by minimizing the cost $J_{\pi}(\theta)$, which consists of the KL divergence to the optimal policy $\pi^* (a_t|s_t)$:
\begin{equation}
     J_\pi(\theta) = \mathbb{E}_{\rho_\pi} [J_\pi(\theta; s_\tau)] = \mathbb{E}_{\rho_\pi} \left[ \mathrm{KL} \left( \pi_{\theta}(a_\tau|s_\tau) \left\Vert \frac{\exp ( Q(s_\tau, a_\tau)/ \kappa)}{Z} \right. \right) \right].
     \label{eq:kl_div}
\end{equation}

This KL objective cannot be analytically computed because $\pi_{\theta}$ and $Z$ are intractable. Therefore, similarly to~\cite{Haarnoja2017reinforcementlearningwithdeepenergybasedpolicy}, we optimize Eq.~\eqref{eq:kl_div} via approximate sampling, and Stein Variational Gradient Descent (SVGD)~\cite{liu2016steinvaritaionalgradientdescent}. 
The Q network is trained using the target in~\cite{sarsa-lambda}.

SVGD is a particle-based algorithm used to approximate a target probability distribution $p(x)$ by moving a set of particles $x_i\in\mathbb{R}^d\sim q(x)$ through a smooth transformation $T(x_i) = x_i + \epsilon \phi(x_i)$ to match that distribution. When the transformation $\phi(x)$ belongs to a Reproducing Kernel Hilbert Space (RKHS) $\mathcal{H}$, the optimal direction $\phi^*(x)$ minimizing the KL divergence between $p(x)$ and $q(x)$ is~\cite{liu2016kernelizedsteindiscrepancyforgoodnessoffittests}:
\begin{equation}
    \phi^*(x)= \frac{\phi^*_{q,p}(x)}{\Vert\phi^*_{q,p}(x)\Vert_{\mathcal{H}^d}} \quad \phi_{q,p}^*(x) = \mathbb{E}_{x'\sim q}[k(x', x)\nabla_{x'}\log p(x') + \nabla_{x'}k(x', x)],
    \label{eq:svgd-dir}
\end{equation}
where $k(x',x)$ is the positive definite kernel function and $\mathcal{H}^d$ is the $d$-dimensional norm that uniquely determine $\mathcal{H}$. 

We can use Eq.~\eqref{eq:svgd-dir} to optimize the policy loss $J_\pi(\theta)$ in Eq.~\ref{eq:kl_div} by using as random variables the actions $a_t$, as sampling distribution $q(a_t)=\pi_\theta(a_t|s_t)$, and as target distribution $p(a_t)=\text{exp}(Q(s_t,a_t)/\kappa)/Z$. Interestingly, we can set $\partial J_\pi(\theta)/\partial a_t \propto \phi_{q,p}^*(a_t)$, even if, strictly speaking, $\phi_{q,p}^*(a_t)$ is not the gradient of $J_\pi(\theta)$ since $\phi^*(a_t)\in\mathcal{H}$~\cite{wang2016learning}.

If a flow matching policy generated an action sample $a_t$ using only one step, $a_{t,1} = a_{t,0} + v_\theta(a_{t,0}, 0)$ where $a_{t,0} \sim \mathcal{N}(0, I)$, then we could use the following loss to optimize $J_\pi(\theta$):
\begin{equation}
    \frac{\partial J_\pi(\theta)}{\partial \theta} \propto \mathbb{E}_{a_{t,0} \sim \mathcal{N}(0, I)} \left [ \phi^*(a_{t,1}) \frac{\partial a_{t,1}}{\partial \theta} \right ] = \mathbb{E}_{a_{t,0} \sim \mathcal{N}(0, I)} \left [ \phi^*(a_{t,1}) \frac{\partial v_\theta(a_{t,0} , 0)}{\partial \theta} \right ].
    \label{eq:policy_gradient}
\end{equation}

From inspection of Eq.~\eqref{eq:policy_gradient}, we see that this RL formulation aligns the policy’s vector field over the action manifold with $\phi^*(a_t)$. In addition, as shown in Eq. \eqref{eq:svgd-dir}, $\phi^*(a_t)$ is proportional to the critic gradient, drawing a parallel to adjoint matching~\cite{qiyang2026qam} and energy-guided flow-matching~\cite{alles2025flowq}. However, in contrast to these methods, our approach additionally incorporates a repulsive term, $\nabla_{a'}k(a',a)$, which prevents particle collapse. This mechanism fosters diverse exploration and stabilizes the learning process.

From a practical standpoint, this formulation is particularly advantageous for flow-matching policies. It bypasses the need to compute the normalization constant $Z$ and the computationally expensive log-likelihood of the flow-matching policy. Instead, the algorithm remains entirely sample-based, requiring only realizations from $\pi_\theta(a_t|s_t)$.

However, as outlined in Sec.~\ref{sec:flow_match}, obtaining action samples requires multiple iterations of the flow model $v_\theta$. Direct backpropagation over iterations, in particular for large number of steps, could dilute the learning signal. In the next section, we present how our algorithm can naturally overcome this issue thanks to its transport formulation.

\subsection{Reinforcement Learning with Density Transport}
\label{sec:multistep-optimization}

To resolve the gradient dilution problem, we observe that any intermediate prediction $a_{t,\tau}$ (where $\tau < 1$) can be mapped to a corresponding sample $a^\dagger_{t,\tau}$ on the policy's action manifold. This allows us to evaluate the gradient in Eq. \eqref{eq:policy_gradient} across all intermediate steps, bypassing the need for backpropagation through the denoising process.

We define $a^\dagger_{t,\tau}=\mathbb{E}_{p_{\tau:1}}[a_{t,1} \mid a_{t,\tau}]$ as the expected posterior of $a_{t,\tau}$~\cite{yaron2024flowmatchingtutorial}. Tianhong~\etal~\cite{tian2025backtobasics} show:
\begin{equation}
    a^\dagger_{t,\tau} = a_{t,\tau} + (1-\tau) v (a_{t,\tau}, \tau; \theta).
\end{equation}
Let's denote the trajectory of $a^\dagger_{t,\tau}$ moving along the transport field $T(a) = id + \epsilon \phi^*(a)$ as $a^\dagger_{t,\tau}(\epsilon)$, where $\epsilon \in \mathbb{R}$, and the corresponding parameter change required in the network's backbone to predict $a^\dagger_{t,\tau}(\epsilon)$ as $\theta = \theta(\epsilon)$. 
Since $a_{t, \tau}, t, \tau$ are fixed, we can omit these symbols and write $a^\dagger_{t,\tau} = a^\dagger(\epsilon)$, $v (a_{t,\tau}, \tau; \theta) = v(\theta(\epsilon)) = v(\epsilon) = (a^{\dagger}(\epsilon) - a)/(1-\tau)$. 
We can take the derivative on both sides w.r.t $\epsilon$ of the last equation at $\epsilon = 0$ and get 
\begin{equation}
     \left. \frac{ dv(\epsilon)}{d\epsilon} \right|_{\epsilon=0} = \frac{1}{1 -\tau} \left. \frac{da^\dagger (\epsilon)}{d\epsilon} \right|_{\epsilon=0}  = \frac{1}{1 - \tau} \phi^*(a^\dagger),
     \label{eq:dv}
\end{equation}
where the latest step is justified by the definition of the transport field $T(a)$. 

We now need to find which update in the parameters $\xi=\frac{\delta\theta}{\delta\epsilon}$ will align the flow field to $T$ at $\epsilon=0$. To do so, we can use the chain rule to write:
\begin{equation}
   \frac{dv(\epsilon)}{d\epsilon} = \frac{\delta v}{\delta \theta}\frac{\delta\theta}{\delta\epsilon} = \frac{\delta v}{\delta\theta} \xi.
   \label{eq:dtheta}
\end{equation}
From Eq.~\eqref{eq:dv} and Eq.~\eqref{eq:dtheta}, one can see that the corresponding $\xi$ 
should minimize:
\begin{equation}
    \xi^*  = \arg \min \limits_{\xi} \left\Vert \frac{\delta v}{\delta\theta}
    \xi - \frac{1}{1 - \tau} \phi^*(a^\dagger) \right \Vert^2
\end{equation}
Even though computing the exact update requires matrix inversion, we can still do a single gradient descent step~\cite{wang2016learning} and compute the approximation $\hat{\xi}$ as:
\begin{equation}
    \hat{\xi} = \frac{1}{(1 - \tau)} \frac{\delta v}{\delta \theta}^\top \phi^*(a^\dagger) = \left [ \partial_\theta \left(\frac{v_\theta^\top \phi^*(\mathrm{sg}(a^\dagger)) }{(1 - \tau) }  \right) \right]
\end{equation}
where $\mathrm{sg}(\cdot)$ is the stop gradient operator. 
In summary, this justifies the following RL loss:
\begin{equation}
    \mathcal{L}_{\mathrm{RL}} = \mathbb{E}_{\tau\sim \mathcal{U}[0, 1], a_{t,\tau} \sim p_\tau(\cdot|s_t)} \left[ \frac{v_\theta(a_{t,\tau}, \tau)^\top \phi^*(\mathrm{sg}(a^\dagger_{t,\tau}))}{T(1 - \tau)} \right]
\end{equation}
where $T$ is the total number of ODE steps.
We add two regularization terms to the training process. 
To regularize the learned flow to be straight and keep the optimal transport property~\cite{Lipman2022FlowMatching}, we add a consistency loss to the model:
\begin{equation}
    \mathcal{L}_{\mathrm{consist}} = \mathbb{E}_{a_{t,\tau} \sim p_\tau(\cdot \mid s_t), a_{t,1}\sim p_1(\cdot \mid s_t)=\pi_\theta(\cdot \mid s_t)} \left[ \Vert a_{t,\tau}^\dagger  - a_{t,1}\Vert ^2 / (T(1 - \tau))^2 \right].
\end{equation} In addition, when finetuning a pretrained flow-matching model, we add a constraint loss $\mathcal{L}_{\mathrm{constraint}}$ to keep the finetuned policy close to the pretrained model.
We choose Fisher Divergence as the constraint because the flow matching policy computes the score function of the distribution~\cite{song2021scorebasedgenerativemodeling}.
\begin{equation}
    \mathcal{L}_{\mathrm{constraint}} = \mathcal{F}(\pi_\theta \Vert \pi_{\bar{\theta}}) = \mathbb{E}_{\tau \sim \mathcal{U}[0, 1], a_{t,\tau} \sim p_\tau(\cdot \mid s_t)} \Vert v_\theta (a_{t,\tau}, \tau) - v_{\bar{\theta}}(a_{t,\tau}, \tau) \Vert^2
\end{equation}

In summary, the complete loss used in flow policy optimization is:
\begin{equation}
    \mathcal{L}_{\mathrm{actor}} =\mathcal{L}_{\mathrm{RL}} + \lambda_1 \mathcal{L}_{\mathrm{consist}} + \lambda_2\mathcal{L}_{\mathrm{constraint}}.
    \label{eq:tfpo_loss}
\end{equation}
The full RLDT algorithm is summarized in the appendix Algorithm~\ref {alg:pipeline}.




\section{Experiments} 
We conduct experiments on a variety of environment settings, including dense reward (OpenAI Gym), long-horizon robot manipulation with sparse rewards (Furniture Bench), and vision-based robot manipulation tasks (Robomimic). For benchmarking, we adopt the same environment setting below for all the methods we test: 

\textbf{OpenAI Gym.} We finetune the flow matching policy using RLDT on the "Walker2d", "Hopper", and "HalfCheetah" tasks. 
We use the original dense reward setting during the finetuning stage.
The policy takes the state as input and predicts action with chunk size $T_a = 4$.

\textbf{Furniture-Bench.} We show that our method could also improve the base policy in long-horizon and sparse reward multistage tasks in the Furniture-Bench environment~\cite{heo2023furniturebench}. 
We chose the \textit{One-Leg} and \textit{Lamp} settings with Low and Medium randomness for the experiment. The \textit{One-Leg} task has a single stage, and the \textit{Lamp} task has two stages. 
The robot will get a single reward at the time when it successfully completes one stage.
The policy takes the state as input and predicts action with chunk size $T_a = 8$.

\textbf{Robomimic.} To prove the effectiveness of our method when the flow-policy operates on high-dimensional visual input, we run experiments in the robomimic environment~\cite{robomimic2021}. Specifically, we chose the \textit{Square} and \textit{Transport} tasks. Our policy takes the proprioception and RGB image as input and predicts action with chunk size $T_a = 4$ for Square and $T_a = 8$ for Transport.

\subsection{Implementation Details}
For the base policy, we adopt the same network architecture as DPPO. 
For a fair comparison, we train all base policies on the same demonstrations as DPPO, and use the same learning rate and training steps. 
The only difference is the training objective, where DPPO uses the DDPM objective while we use the Flow Matching objective. 
The ODE step $T$ is 8 for the OpenAI Gym and Robomimic setup and 32 for the Furniture Bench setup.

For each state sample $s_\tau$, we generate $K=8$ action samples for SVGD computation.
We choose the temperature $\kappa$ as $\kappa = \frac{\kappa_0}{K} \sum_{m=1}^K \Vert \nabla_aQ (s, a_m) \Vert $. 
We set $\kappa_0 = 0.001$. 
We choose the RBF kernel as the kernel function for $\phi^*(x)$, where $k(x, x') = \exp(- \Vert x - x' \Vert^2 / (2h^2))$. 
The radius $h$ is the median pairwise distance of the $K$ action samples.
We use the $K$ clean generated samples to construct the transport field $\phi^*(a)$.
To compute the SVGD loss, for each sample in the $K$ ensemble, we uniformly randomly select a timestep $t_i = i / T$ for $i \in \{0, 1, ..., T - 1\}$ as the intermediate $a_{t_i}$.
The loss weights are $\lambda_1 = 0.5$ and $\lambda_2 = 0.2$. 

For state-based tasks, we use fully connected networks as the Q network.
We choose SiLU as the activation function. 
For vision-based tasks, we use the Transformer as the feature encoder to extract features from the images and robot proprioception.
Afterwards, we concatenate the feature with the action and pass it to a fully connected network to estimate the Q value.  
The details for the network architecture and hyperparameters for Q learning are presented in the appendix. 

\subsection{Results}

We design our experimental setting to answer the following set of questions. We present robustness to hyperparameters (entropy coefficient $\kappa_0$ and number of particles $K$) in the appendix.

\paragraph{Q1: How effective is RLDT compared to state-of-the-art methods?}
We compare our method against prior work that approximate the log-likelihood with different lower bounds. Specifically, we select \textbf{DPPO}~\cite{rendiffusion}, \textbf{ReinFlow}~\cite{zhang2025reinflowfinetuningflowmatching} and \textbf{FPO++}~\cite{yi2026fpo++}. We additionally compare to \textbf{QAM}~\cite{qiyang2026qam}, which generates targets for intermediate denoising steps using adjoint matching.

\begin{figure}
    \centering
    \includegraphics[width=1.0\linewidth]{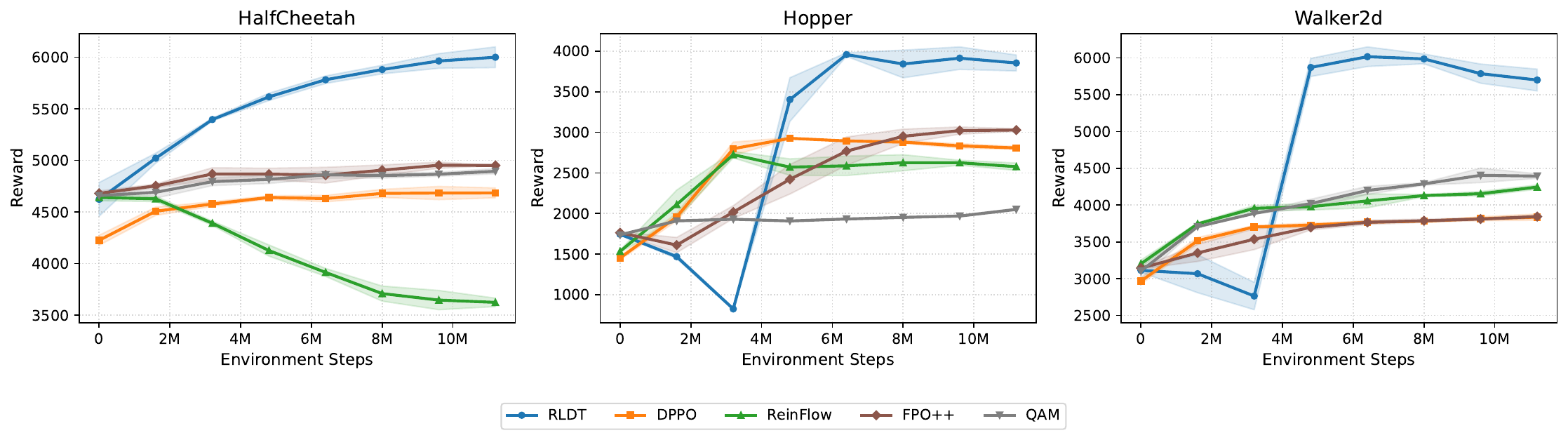}
    \caption{\textbf{Dense Reward Tasks} 
            We run each setting for 3 different seeds and plot the mean and standard deviation.}
    \label{fig:likelihood-gym-benchmark}
    \vspace{-3mm}
\end{figure}

\begin{figure}
    \centering
    \includegraphics[width=1.0\linewidth]{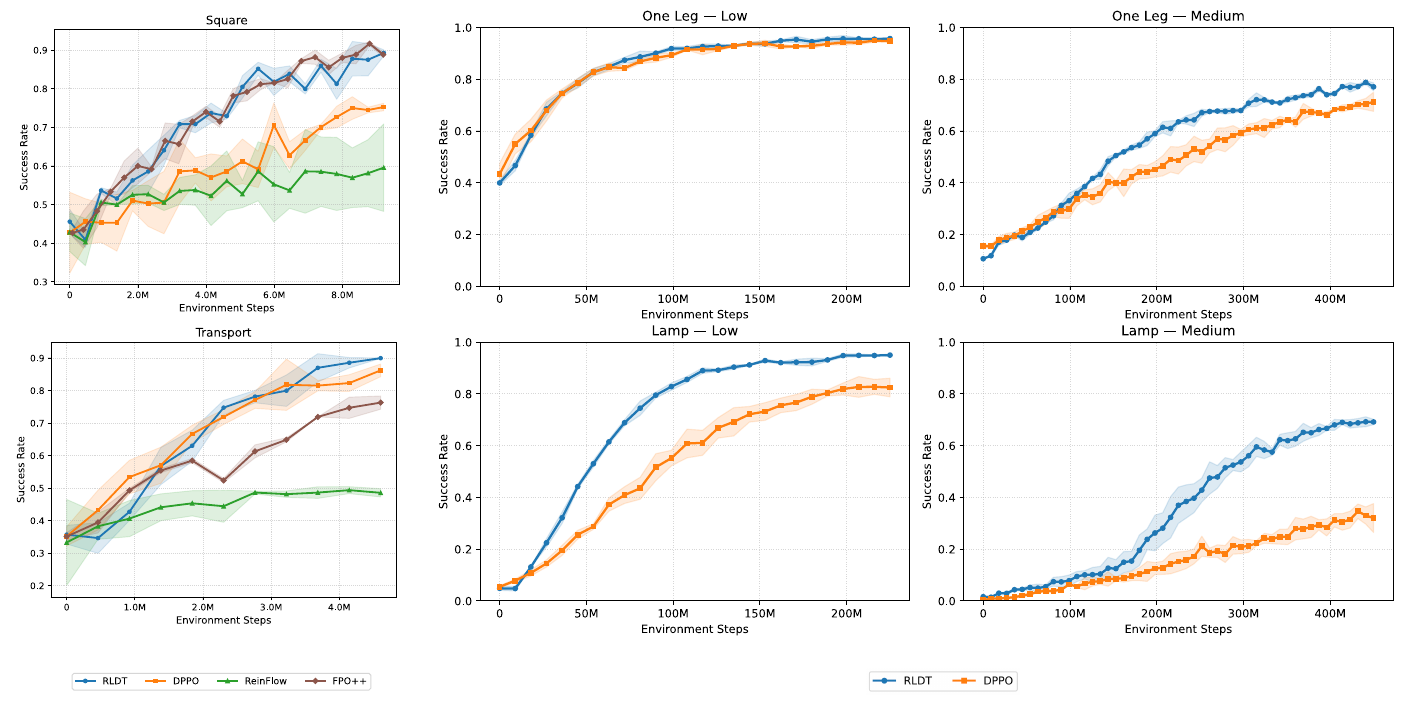}
    \caption{\textbf{Sparse Reward Tasks} The left plot presents results on the visual-based robomimic benchmark, and the right plot on state-based furniture bench tasks. Results are averaged over 3 seeds. }
    \label{fig:likelihood-robomimic-benchmark}
\end{figure}

\textbf{Baselines Implementation.} DPPO and ReinFlow approximate the log-likelihood with the joint probability of all latent variables. Conversely, FPO++ approximates the log-likelihood using the ELBO lower bound.
For DPPO and ReinFlow, we use the official implementations and parameters.
For FPO++, we use the same pre-trained flow-matching policy as RLDT. 
We sample 16 time steps to compute the conditional flow matching loss for the ratio computation.
We compute the ratio for each individual sample. 
We use the ASPO setting for clipping, where the ratios are clipped using the PPO~\cite{schulman2017ppo} method for positive advantages, and the ratios are clipped using the SPO~\cite{spo} for negative advantages.
We sweep the clipratio parameter for FPO++ of \{0.01, 0.02, 0.04\} and find 0.01 works the best. For QAM, we keep the Q network learning parameters the same as RLDT and only change the policy update scheme.
For the denoising steps, we use the same number of denoising steps as RLDT. For QAM, the slow actor $\pi_\beta$ is fixed to the pre-trained flow matching policy.
We sweep the inverse temperature parameter over \{0.1, 0.3, 0.05\} for the inverse temperature in QAM and find 0.3 has the best performance, which is also recommended by the paper.

Results on the dense-reward Gym benchmarks are summarized in Fig.~\ref{fig:likelihood-gym-benchmark}.
RLDT achieves the highest rewards across all baselines by a significant margin.
DPPO results match those reported in the original paper, while ReinFlow underperforms on HalfCheetah and Hopper despite using their official code. QAM performance varies across tasks, in some outperforming other baselines and in others underperforming them. Nonetheless, it always underperformed our method.

Results on the robot manipulation tasks with sparse rewards are summarized in Fig.~\ref{fig:likelihood-robomimic-benchmark}. In these benchmarks, we don't compare to QAM, since the method was not tested on such robotics tasks, and we found it challenging to tune to a reasonable performance.

We start from the robomimic benchmark (Fig.~\ref{fig:likelihood-robomimic-benchmark}-left), where the policy has image-based inputs. In the \textit{Square} task, RLDT starts with a higher initial success rate and continues to improve, achieving a final success rate of approximately $90\%$. 
For the \textit{Transport} task, RLDT improves slightly more slowly than DPPO early in training but converges to comparable performance.
FPO++'s performance is close to RLDT on the \textit{Square} task, while it achieves a lower final success rate around 75\% on the \textit{Transport} task.
As in the Gym benchmark, we find Reinflow to underperform other methods,  especially in the \textit{Transport} task.

We conclude with FurnitureBench tasks (Fig.~\ref{fig:likelihood-robomimic-benchmark}-right), where the policy has state and proprioception as input. Here, we only compare to the strongest baseline, DPPO. RLDT achieves performance comparable to it on the \textit{One Leg Low} and \textit{One Leg Med} tasks.
However, on the \textit{Lamp Med} task, RLDT reaches a success rate of approximately 70\%, whereas DPPO achieves only 30\%.


\begin{figure}
    \centering
    \includegraphics[width=1.0\linewidth]{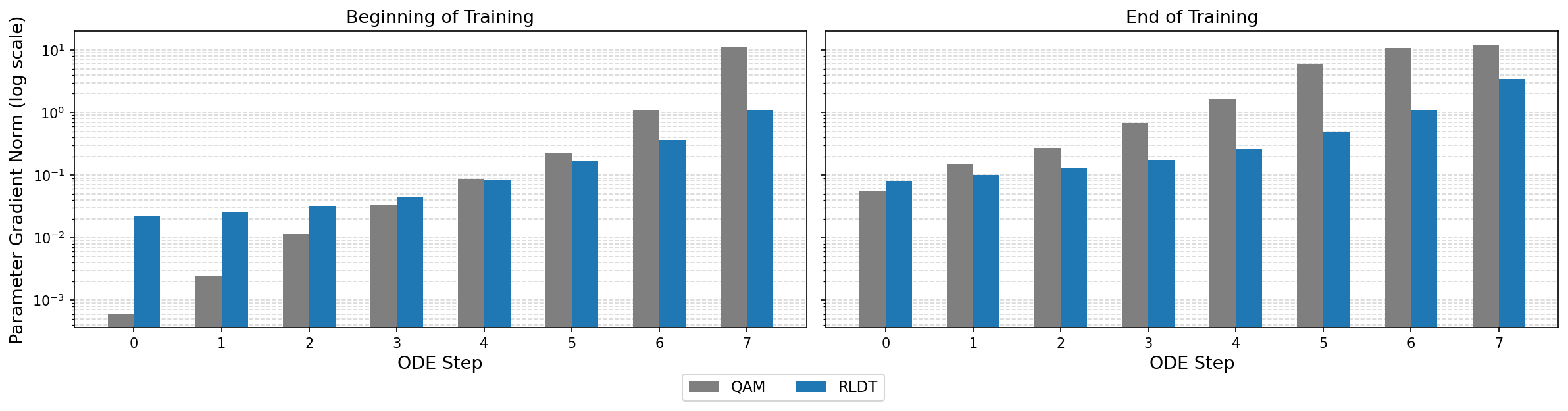}
    \caption{\textbf{Gradient Norm for ODE Timestep} 
    Left: Parameter gradient norm computed using the pre-trained policy at the beginning of finetuning. 
    Right: Parameter gradient norm computed using the finetuned policy at the end of finetuning. 
    0 is the first ODE step and 7 timestep is the last ODE step.}
    \label{fig:gradient-magnitude}
    \vspace{-4mm}
\end{figure}

\paragraph{Q2: Does RLDT balance the learning signal across denoising timesteps?} 
To analyze the effectiveness of the proposed optimization method, we compare parameter gradient magnitudes across ODE timesteps to QAM~\cite{qiyang2026qam}. We plot the gradient norm at the beginning and end of the training for the \textit{Hopper} environment in Fig.~\ref{fig:gradient-magnitude}.

At the beginning of training, QAM has a very unbalanced gradient from different ODE steps. 
The initial step contributes less than 0.001 to the parameter update, indicating that the early ODE step output does not contribute to the overall gradient.
At the end of training, this problem is mitigated because the finetuned policy deviates from the reference policy, which contributes to a larger gradient at ODE step 0.
For RLDT, the gradient contribution from each ODE is more evenly distributed, and the distribution remains steady over the training process. This empirically leads to faster and more stable training.

\begin{figure}
    \centering
    \includegraphics[width=\linewidth]{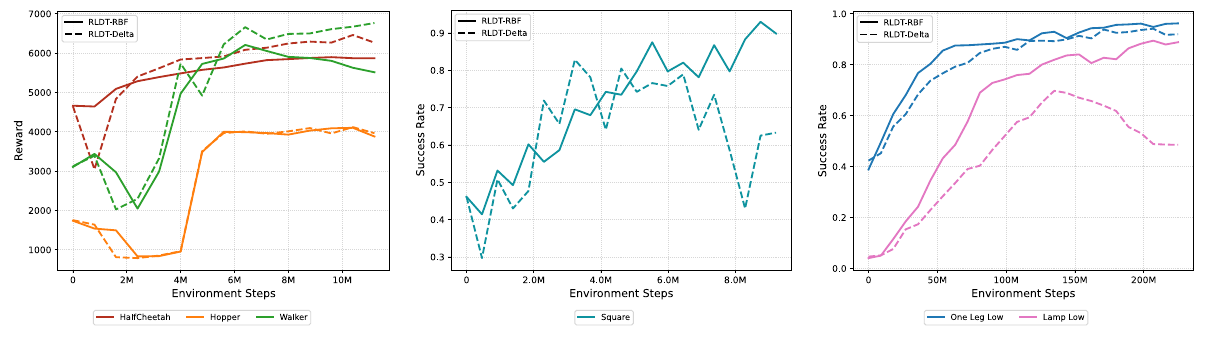}
    \caption{\textbf{Ablation on Kernel Function} 
    We compare the RBF kernel (RLDT-RBF) against the Delta kernel (RLDT-Delta) in the SVGD computation.}
    \label{fig:kernel-ablation-furniture}
    \vspace{-4mm}
\end{figure}


\paragraph{Q3: What is the role of the kernel function in the transport field?} 
To analyze the impact of the RBF kernel, we compare with \textbf{RLDT-Delta}, which uses a delta function for the kernel function. 
Specifically, $k(x_i, x_j) = 1$ if $\Vert x_j  - x_i\Vert = 0$ and $k(x_i, x_j) = 0$ if $\Vert x_j  - x_i\Vert \neq 0$.
In this case, we get $\phi^*(\mathrm{sg}(a^\dagger_t)) = \nabla_a Q(s, a_t^\dagger)$, i.e., the critic's gradient. Therefore, RLDT-Delta does not have repulsive forces between particles.
We run experiments on the Gym and the Robomimic environment.

The result are summarized in Fig.~\ref{fig:kernel-ablation-furniture}.
On the Gym task, the two kernels behave similarly, with RLDT-Delta slightly outperforming RLDT-RBF. This is expected since exploration is simple in these dense reward tasks: following the critic's gradient is sufficient for high performance. We confirm this by checking the norm $\Vert \phi^*(a_t^\dagger)\Vert$, and find that \textbf{RLDT-Delta} always has a higher norm than \textbf{RLDT-RBF} and the $\mathcal{L}_{\mathrm{constraint}}$ is consistently higher, indicating a more drastic update of the actor (Figures in the Appendix). 
However, this behavior changes when moving to sparse reward robotics tasks. 
For the \textit{Square} task and the \textit{Lamp-Low} task, \textbf{RLDT-Delta} fails to improve the policy after half of the training. Interestingly, the performance drops near the end, potentially due to an overfitted critic. Overall, this shows that the delta kernel leads to instabilities for sparse reward tasks.

While the relative trends between these kernels were expected, we unexpectedly found that the delta function does not lead to mode collapse, i.e., all particles producing similar actions. Conversely, we found that the pair-wise distance between action samples is positive for all tasks. We hypothesize that this is because the Q function keeps evolving, so that the samples cannot stop in the same location. However, only relying on the Q function for exploration is not sufficient for tasks with sparse rewards.



\section{Conclusion}

We presented RLDT, a reinforcement learning finetuning framework for flow-matching policies that formulates policy improvement as probability transport on the action manifold.
Experiments across dense-reward continuous control (OpenAI Gym), sparse-reward long-horizon manipulation (Furniture-Bench), and visual manipulation (Robomimic) demonstrate that RLDT consistently outperforms state-of-the-art baselines.
%
Despite these advantages, RLDT introduces a computational overhead due to the parallel simulation and gradient computation required for SVGD particles. However, our ablation (see Appendix) shows that only four particles are sufficient for high performance. While this requirement may pose challenges for real-robot RL under a limited computation budget, it is well-suited to simulation-to-real transfer pipelines~\cite{yi2026fpo++}.
Overall, our work demonstrates the advantages of formulating RL as the alignment of a flow-matching transport field with a target field that drives action probabilities toward the optimal policy density.


\clearpage
{
\small
\bibliography{references}
}



\newpage
\appendix 

\section{Algorithm Summary}
Our method is summarized in Algorithm~\ref{alg:pipeline}.

\begin{algorithm}[ht]
\caption{RLDT}
\label{alg:pipeline}
Flow matching policy parameters $\theta$, Q-network parameters $\{\phi_i\}, i=1,2$\;
Assign target parameters: $\bar{\theta} \leftarrow \theta,\ \bar{\phi}_i \leftarrow \phi_i$\;
$\mathcal{D} \leftarrow$ empty replay buffer\;
\For{each training iteration}{
    \tcp{Environment Rollout}
    \For{$m = 1$ \KwTo $M$}{
        Sample $a_\tau \sim \pi_\theta(a_\tau \mid s_\tau)$ from the flow matching policy\;
        Execute $a_\tau$, observe $(s_\tau, a_\tau, r_\tau, s_{\tau+1})$; store transition in $\mathcal{D}$\;
    }
    Sample minibatch $(s_\tau, a_\tau, r_\tau, s_{\tau+1})$ from $\mathcal{D}$\;
    \tcp{Q-Network Update}
    \For{each gradient step}{
        Compute the Q target using Eq.(5) from~\cite{sarsa-lambda}\;
        Update $\phi_i$ by minimizing $\|Q_{\phi_i}(s_\tau, a_\tau) - y_i\|^2$ for $i=1,2$\;
        Update target networks: $\bar{\phi}_i \leftarrow \rho\,\phi_i + (1-\rho)\,\bar{\phi}_i$ for $i=1,2$\;
    }
    \tcp{Policy Update via RLDT}
    \For{each gradient step}{
        Sample $\tau \sim \mathcal{U}[0,1]$, $a_{t,0} \sim \mathcal{N}(0,I)$\; Run ODE in Eq.\eqref{eq:flow-ode} to get intermediate $a_t$\;
        Compute manifold prediction: $a^\dagger_{t,\tau} \leftarrow a_{t,\tau} + (1-\tau)\,v_\theta(a_{t,\tau}, \tau \mid s_t)$\;
        Compute transport direction $\phi^*(a^\dagger_{t,\tau})$ via Eq.~\eqref{eq:svgd-dir}\;
        Update $\theta$ by minimizing Eq.~\eqref{eq:tfpo_loss};  
    }
    Update target $\bar{\theta}$ by $\bar{\theta} \leftarrow \theta$\;
}
\end{algorithm}

\section{Training Parameters}

\subsection{Base Policy Training}

All base policies are trained via behavioral cloning using the flow matching objective in Eq.~\eqref{eq:flow_train} with a cosine-annealing learning rate scheduler.
The differences among the three environments are listed below:

\paragraph{OpenAI Gym.}
The policy is a three-layer residual MLP (hidden dim 512, ReLU activations, time embedding dim 16, no layer normalization) conditioned on a single observation frame.
Training runs for 200 epochs with batch size 128, initial learning rate $10^{-3}$ decaying to $10^{-4}$, and a single warm-up step.

\paragraph{FurnitureBench.}
The higher observation dimensionality ($d_s = 58$, $d_a = 10$) calls for a deeper network: a seven-layer residual MLP with hidden dim 1024, a two-layer conditioning MLP (dims [512, 64]), time embedding dim 32, and layer normalization.
Training runs for 8\,000 epochs with batch size 256, initial learning rate $10^{-4}$ decaying to $10^{-5}$, and 100 warm-up steps.

\paragraph{Robomimic.}
The visual policy uses a Vision Transformer (ViT) image encoder (patch size 8, depth 1, embedding dim 128, 4 attention heads, spatial embedding dim 128) followed by a three-layer residual MLP (hidden dim 768, time embedding dim 32) with online data augmentation.
Proprioceptive observations ($d_s = 9$ for \textit{Square}, $d_s = 18$ for \textit{Transport}) and RGB images ($3 \times 96 \times 96$) are fused via the spatial embedding.
Training runs for 2\,000 epochs with batch size 256, initial learning rate $10^{-4}$ decaying to $10^{-5}$, and 100 warm-up steps.

A summary of the key hyperparameters is provided in Table~\ref{tab:pretrain-params}.

\begin{table}[h]
\centering
\caption{Base policy pretraining hyperparameters.}
\label{tab:pretrain-params}
\resizebox{\textwidth}{!}{
\begin{tabular}{lcccccc}
\toprule
\textbf{Environment} & \textbf{Network} & \textbf{Hidden dim} & \textbf{$T_a$} & \textbf{Epochs} & \textbf{Batch size} & \textbf{LR} \\
\midrule
Gym (HalfCheetah, Hopper, Walker2d) & MLP $\times 3$ & 512 & 4 & 200 & 128 & $10^{-3} \to 10^{-4}$ \\
FurnitureBench & MLP $\times 7$ & 1024 & 8 & 8000 & 256 & $10^{-4} \to 10^{-5}$ \\
Robomimic (\textit{Square}) & ViT + MLP $\times 3$ & 768 & 4 & 8000 & 256 & $10^{-4} \to 10^{-5}$ \\
Robomimic (\textit{Transport}) & ViT + MLP $\times 3$ & 768 & 8 & 8000 & 256 & $10^{-4} \to 10^{-5}$ \\
\bottomrule
\end{tabular}
}
\end{table}

\subsection{RL Training}

All environments share the same SVGD and actor update settings: temperature $\kappa_0 = 0.001$, $K = 8$ particles, actor learning rate $2 \times 10^{-5}$ (32-step warmup), and 32 policy gradient steps per iteration.
The constraint coefficient is $\lambda_2 = 0.2$ and the RBF kernel bandwidth is set to the median pairwise distance of the $K$ action samples.
The critic uses a double-Q ensemble ($M = 2$) at learning rate $10^{-3}$.
Two critic warm-up iterations are performed before the first actor update for FurnitureBench and Robomimic.

The three environment families differ in discount factor, number of training iterations, Q-update steps per iteration, critic architecture, consistency loss weight $\lambda_1$, and reward handling, as detailed below.

\paragraph{OpenAI Gym.}
The critic is a three-layer MLP (hidden dim 256, ReLU, no residual) taking the concatenated state and action chunk as input.
Rewards are normalized online with EMA (smoothing $\alpha = 0.01$).
We smooth over the average of rewards per step.
Training runs for 70 iterations with 100 parallel environments, Q-network updated for 1\,024 gradient steps per iteration, $\gamma = 0.99$, Q target momentum $\rho = 0.998$, and $\lambda = 0.0$. The training takes around 1.5 GPU hours on a single NVIDIA 2080Ti GPU.

\paragraph{FurnitureBench.}
The critic is a three-layer residual MLP (hidden dim 512, SiLU, layer normalization disabled).
Rewards are sparse and fixed to $[0, 1]$ per stage; balanced sampling (equal nonzero and zero reward trajectories) is used to mitigate reward sparsity.
Training runs for 500 iterations for medium randomness and 250 iterations for low randomness with 1\,000 parallel environments, Q-network updated for 2\,048 gradient steps per iteration, $\gamma = 0.995$, Q target momentum $\rho = 0.998$, and $\lambda = 0.5$. The training takes around 48 GPU hours on a single NVIDIA A40 GPU.

\paragraph{Robomimic.}
Both the actor and critic share the same ViT backbone (patch size 8, depth 1, embedding dim 128, 4 heads) to encode RGB observations; the critic head is a three-layer MLP (hidden dim 256, SiLU).
Sparse binary rewards are used with balanced sampling and Q-target clipping to $[0, 1]$.
Training runs for 200 iterations with 64 parallel environments ($n_\text{steps} = 200$ per iteration), Q-network updated for 512 gradient steps per iteration, $\gamma = 0.999$, Q target momentum $\rho = 0.995$, and $\lambda = 0.5$.The training takes around 30 GPU hours on a single NVIDIA A40 GPU.

A summary of key hyperparameters of RL training is provided in Table~\ref{tab:rl-params}.

\begin{table}[h]
\centering
\caption{RL fine-tuning hyperparameters.}
\label{tab:rl-params}
\resizebox{\textwidth}{!}{
\begin{tabular}{lcccccccc}
\toprule
\textbf{Environment} & \textbf{Iterations} & \textbf{$n_\text{envs}$} & \textbf{Q steps/itr} & \textbf{$\gamma$} & \textbf{$\rho$} & \textbf{$\lambda$} & \textbf{Critic} & \textbf{Reward} \\
\midrule
Gym          & 70  & 100  & 1024 & 0.990 & 0.998 & 0.0  & MLP $\times 3$ (256) & EMA norm. \\
FurnitureBench & 500 & 1000 & 2048 & 0.995 & 0.998 & 0.5 & MLP $\times 3$ (512) & Sparse $[0,1]$ \\
Robomimic    & 200 & 64   & 512  & 0.999 & 0.995 & 0.5  & ViT + MLP $\times 3$ (256) & Sparse $[0,1]$ \\
\bottomrule
\end{tabular}
}
\end{table}


\section{More Ablation Study}

\begin{figure}[ht]
    \centering
    \includegraphics[width=1.0\linewidth]{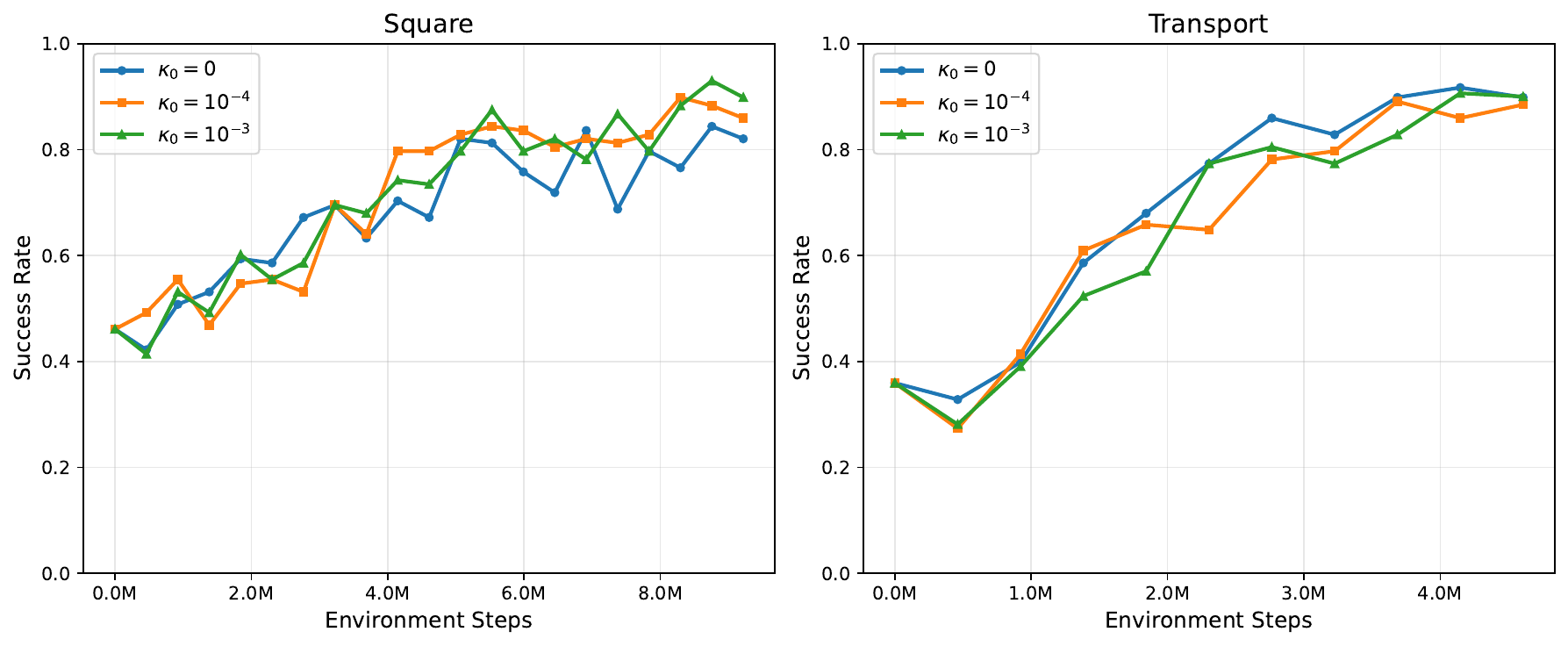}
    \caption{\textbf{Ablation Study on Temperature $\kappa_0$ on Gym Task}}
    \label{fig:hyperparameter-temp-gym}
\end{figure}

\subsection{Ablation on Temperature $\kappa_0$}
We analyze two hyperparameters relevant to our methods. 
The first one is the temperature $\kappa_0$. 
We choose three different settings for $\kappa_0$ as \{1e-3, 1e-4, 0\}.
We run the experiment in the Robomimic environment.
The result is summarized in Fig.~\ref{fig:hyperparameter-temp-gym}.
For the \textit{Transport} task, all three temperature settings have similar performance at the end.
For the \textit{Square} task, a higher temperature leads to a higher terminal success rate. 
We attribute this to the exploration led by a higher temperature. 

\begin{figure}
    \centering
    \includegraphics[width=1.0\linewidth]{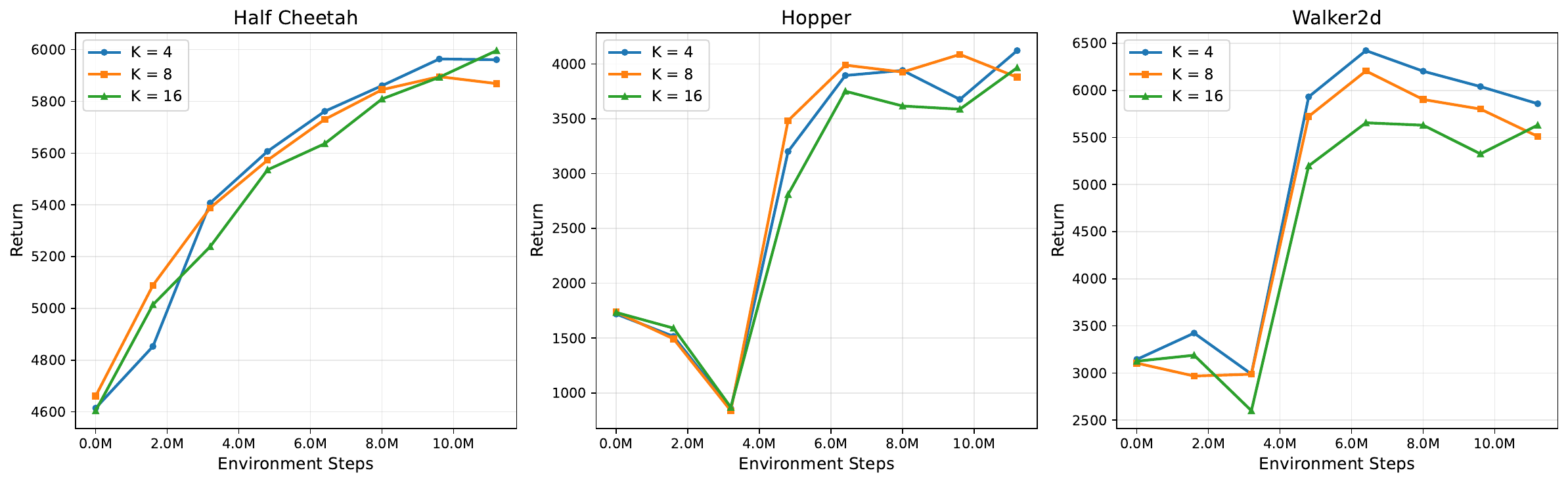}
    \caption{\textbf{Ablation Study on Ensemble Size on Gym Task}}
    \label{fig:hyperparameter-ensemble-gym}
\end{figure}

\subsection{Ablation on ensemble size $K$} 
The second one is the sampled ensemble size $K$ to compute the transport field $\phi^*(a)$.
We test three settings: $K=4,8,16$.
As we can see from Fig.~\ref{fig:hyperparameter-ensemble-gym}, our RLDT has similar performance for different ensemble sizes, indicating our algorithm is not sensitive to this parameter in the Gym task.


\begin{figure}
    \centering
    \includegraphics[width=\linewidth]{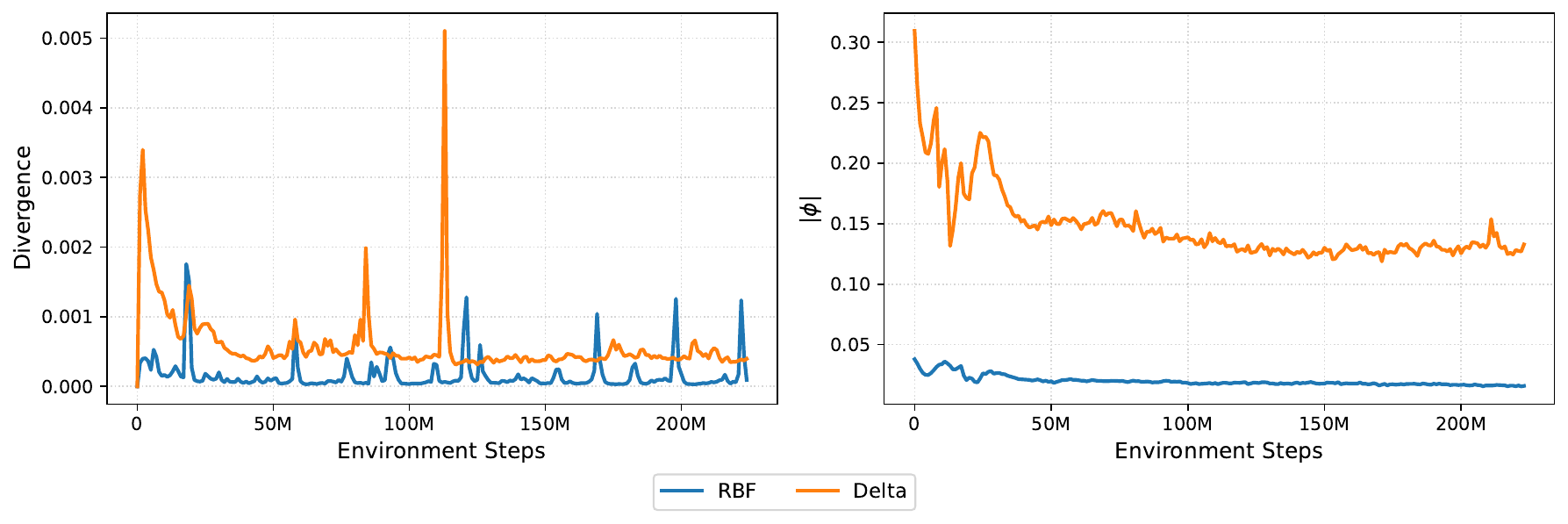}
    \caption{\textbf{Training Statistics for Different Kernel Function} The training statistics on the \textit{Lamp-Low} task. Left: The constraint loss over the training steps. Right: The average norm of $\phi(a^\dagger)$ over the training steps}
    \label{fig:delta-ablation-stats}
\end{figure}

\subsection{More Results for Kernel Function}

To show the difference of delta kernel and RBF kernel. We plot the constraint loss and the average norm of transport velocity $\phi(a^\dagger)$ over the training steps.
The result is summarized in Fig.~\ref{fig:delta-ablation-stats}.
The delta kernel incurs a higher velocity norm than the RBF kernel. 
This is because the RBF kernel smooths the Q gradient value at all the clean samples, while the delta kernel only uses the gradient at a single point.
The higher norm leads to higher constraint loss, indicating the policy updates more drastically using the delta kernel.
There are sharp spikes during training using a delta kernel, indicating the instability incurred by the delta function.

\clearpage

\end{document}